\pdfoutput=1
\documentclass[11pt]{article}
\usepackage[final]{acl}
\usepackage{graphicx}
\usepackage{graphics} 
\usepackage{epsfig}
\usepackage{booktabs}
\usepackage{times}
\usepackage{latexsym}
\usepackage{bbding}
\usepackage{amssymb}
\usepackage{amsmath,amsthm,amssymb,amsfonts}
\usepackage{makecell}
\usepackage{multirow}
\usepackage{xcolor}
\usepackage{colortbl}  
\usepackage{hyperref}
\usepackage{booktabs} 

\usepackage[T1]{fontenc}
\usepackage[utf8]{inputenc}

\usepackage{microtype}

\usepackage{inconsolata}

\title{HyperMoE: Towards Better Mixture of Experts via Transferring Among Experts}

\newcommand*\samethanks[1][\value{footnote}]{\footnotemark[#1]}

\author{
\small
    Hao Zhao\textsuperscript{1}\quad 
    Zihan Qiu\textsuperscript{3}\quad 
    Huijia Wu\textsuperscript{1}\quad 
    Zili Wang\textsuperscript{4}\quad 
    Zhaofeng He\textsuperscript{1}\begin{NoHyper}\thanks{~~Corresponding authors.}\end{NoHyper}\quad 
    Jie Fu\textsuperscript{2}\samethanks[1]
\\
\small
    \textsuperscript{1}Beijing University of Posts and Telecommunications\quad
\textsuperscript{2}Hong Kong University of Science and Technology\\
\small
    \textsuperscript{3}Tsinghua University \quad \textsuperscript{4}INF Technology\\
\\
\small
    \texttt{\{haozhao,huijiawu,zhaofenghe\}@bupt.edu.cn} \quad \texttt{jiefu@ust.hk} \\
\small
    \texttt{qzh11628@gmail.com} \quad \texttt{ziliwang.do.gmail.com}\\
}

\begin{document}
\maketitle
\begin{abstract}
% without significantly increasing computational cost.\
The Mixture of Experts (MoE) for language models has been proven effective in augmenting the capacity of models by dynamically routing each input token to a specific subset of experts for processing. 
Despite the success, most existing methods face a challenge for balance between sparsity and the availability of expert knowledge: enhancing performance through increased use of expert knowledge often results in diminishing sparsity during expert selection. 
To mitigate this contradiction, we propose HyperMoE, a novel MoE framework built upon Hypernetworks. 
This framework integrates the computational processes of MoE with the concept of knowledge transferring in multi-task learning. 
Specific modules generated based on the information of unselected experts serve as supplementary information, which allows the knowledge of experts not selected to be used while maintaining selection sparsity. 
Our comprehensive empirical evaluations across multiple datasets and backbones establish that HyperMoE significantly outperforms existing MoE methods under identical conditions concerning the number of experts. 
Our code is publicly available at \url{https://github.com/Bumble666/Hyper_MoE}

\end{abstract}

\section{Introduction}

The accelerated advancement of large language models has culminated in their widespread application across various domains, including healthcare, education, and social interactions~\citep{NEURIPS2020brown,achiam2023gpt,touvron2023llama}. The remarkable capabilities of these models are attributed to the enhancements in their scale. Nevertheless, the scaling of dense models is often hampered by significant computational demands, posing a challenge to developing the Natural Language Processing (NLP) community. 
In response, sparse activation models have emerged as a solution~\citep{artetxe-etal-2022-efficient,pmlr-v162-du22c}, activating only a subset of parameters for different inputs, thus mitigating computational costs. 
One of the most representative methods is the Mixture of Experts (MoE,~\citet{shazeer2017}), which routers different inputs to specific groups of experts, thereby enlarging the model's capacity without increasing computational burdens.

\begin{figure}
\centering
\includegraphics[width=1.0\columnwidth]{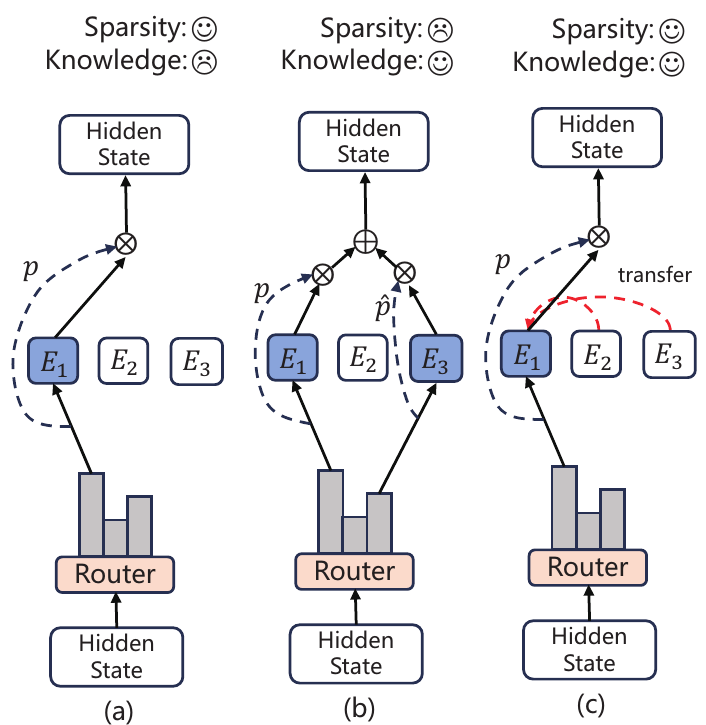}
\caption{A trade-off in MoE: (a) A small number of selectable experts can maintain sparsity but limits the availability of expert knowledge. (b) Increasing the number of selectable experts can improve performance but decrease sparsity. (c) Transferring partial knowledge from the unselected experts $E_{2,3}$ to the selected experts $E_1$ can improve the availability of expert knowledge while maintaining sparsity.}
\label{fig1}
\end{figure}

The key to effectively reducing computational costs lies in the sparsity of expert selection, with the number of experts selected for each token being kept at a lower level. 
In practical applications or experiments, existing works~\citep{NEURIPS2021_92bf5e62,fedus2022switch, pmlr-v162-rajbhandari22a,openmoe2023} usually select only one or two experts per input. 
However, increasing the number of selected experts per token can enhance the availability of expert knowledge and improve the performance of downstream tasks~\cite{NEURIPS2019_f2201f51,shazeer2017,he-etal-2023-merging}. 
This scenario positions MoE model in a predicament akin to a zero-sum game: a choice between increasing the number of available experts to improve performance or preserving a lower level of available experts to ensure sparsity, as depicted in Figure~\ref{fig1}.

To mitigate this contradiction, one solution would be to use the knowledge of other experts to assist the sparsely selected experts. 
This is similar to multi-task learning, which transfers knowledge among related tasks. 
Some works~\citep{karimi-mahabadi-etal-2021-parameter,ivison-peters-2022-hyperdecoders,zhao-etal-2023-prototype} suggest using hypernetworks~\citep{ha2017hypernetworks} to generate task-specific knowledge to enhance positive transfer between tasks. 
Inspired by this, we aim to increase the availability of expert knowledge by transferring the knowledge of unselected experts while sparsely selecting experts.

In this paper, we propose \textbf{HyperMoE}, a novel MoE framework built upon hypernetworks, which captures the information from every expert by leveraging expert-shared hypernetwork while achieving positive expert transfer by generating conditioned modules individually. 
We refer to the information as \textit{cross-expert} information.
Specifically, a HyperMoE consists of HyperExperts, which are generated based on the information of unselected experts and serve as supplementary information for selected experts while maintaining sparsity.

We further improve upon this by introducing the concept of \textit{cross-layer} Hypernetworks: A hypernetwork is shared among all transformer layers, which enables information flow among MoEs in different layers. 
This brings additional efficiency in terms of parameters and computational costs: Despite the additional computation, our method only experienced a decrease\footnote{The degree of decline in speed is related to the scale of the Hypernetworks and the bottleneck size in the generated HyperExpert (similar to $r$ in LoRA). For various tasks, these hyperparameters can be dynamically adjusted to control the delay.} of approximately 15\% in training speed and 10\% in inference speed compared to the standard MoE.

We evaluate HyperMoE on 20 representative NLP datasets across diverse tasks: sequence classification, extractive question answering,  summarization, and text generation. 
Extensive experimental results show that HyperMoE outperforms baselines, including Switch Transformer~\citep{fedus2022switch} with MoE architecture.
This demonstrates the effectiveness of our method in transferring knowledge to experts, which increases the utilization of expert knowledge while keeping the number of experts selected at a low level.

To summarise, our core contributions are:
\begin{itemize}

\item We propose a novel HyperMoE architecture with HyperExpert for MoE framework, which resolves the inherent tension between maintaining sparse expert selection and ensuring sufficient expert availability within MoE.
\item HyperMoE outperforms baselines based on Switch Transformer across a diverse set of NLP tasks, confirming our approach's effectiveness.
\item We show the relevance between selection embeddings, which are based on the context of unselected experts, and selected experts, indicating that the selection embeddings effectively encode the information of knowledge that the currently selected experts need.
\end{itemize}

\section{Background}
\subsection{Mixture of Expert}
\label{s21}
A Mixture of Experts (MoE) typically consists of two parts: the gate model $G$ and a set of expert models $E_1, E_2,\cdots, E_N$. The gate model is used to dynamically select and combine the outputs of the expert models based on the input $x$. As a result, each input will be determined by the collective participation of multiple expert models to obtain the output $y$:
\begin{equation}
    y = \sum_{i=1}^{N} G(x)_iE_i(x).
\end{equation}

The gate model $G(\cdot)$ is a Noisy Top-K Network~\cite{shazeer2017} with parameters $\mathrm{W}_g$ and $\mathrm{W}_{noise}$. This gating method introduces adjustable noise and then retains the top-k values as the final output:
\begin{equation}
    \begin{aligned}
G(x)= & \operatorname{TopK}\left(\operatorname{Softmax} \left(x \mathrm{W}_{g}\right.\right. \\
& \left.\left.+\mathcal{N}(0,1) \operatorname{Softplus}\left(x \mathrm{W}_{\text {noise }}\right)\right)\right),
\end{aligned}
\end{equation}
where $\operatorname{TopK(\cdot)}$ denotes selecting the largest K elements.

MoE allows for flexible adjustment of the contribution of expert models in different input scenarios, thereby improving the overall performance and adaptability of the model.
\begin{figure*}
\centering
\includegraphics[width=1.5\columnwidth]{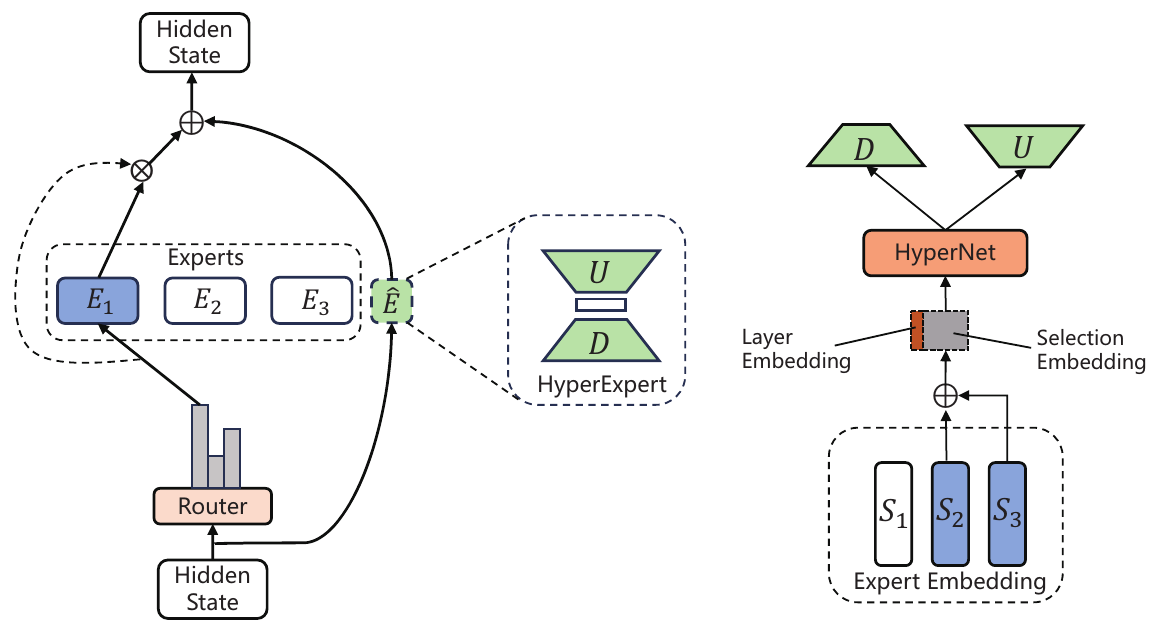}
\caption{Overview of HyperMoE, with a case of one expert is selected. HyperExperts generated from the shared hypernetwork benefit from the cross-expert knowledge within it. Conditional inputs can enhance positive transfer between experts, generating independent modules containing knowledge relevant to the current expert. 
Taking the figure as an example, the selection embedding is obtained by aggregating unselected experts $S_{2,3}$'s embeddings. 
This selection embedding is input into a hypernetwork, which is shared across all experts and all layers, to generate a specific HyperExpert $\hat{E}$ that participates in the computation along with the selected experts $E_1$. The experts $E_2$ and $E_3$ are not activated throughout the process.}
\label{fig2}
\end{figure*}

\subsection{HyperNetworks}
Hypernetwork~\citep{ha2017hypernetworks} can generate parameters to be used by target networks or modules. Specifically, a hypernetwork with independent parameters $\phi$ denoted as $h_\phi $, leverages an context information $z$ to generate the target parameters $\theta$ for the primary network $f_\theta$ and the primary network with an input $x$ is redefined as:
\begin{equation}
    \mathrm{output} =f_\theta(x)=f_{h_\phi(z)}(x).
\end{equation}
This method of flexibly adjusting the parameters of the target network to adapt to different input scenarios is widely used in multi-task learning~\citep{karimi-mahabadi-etal-2021-parameter,ustun-etal-2022-hyper} and few-shot learning~\citep{ponti-etal-2021-parameter}. While generating condition-specific parameters, these parameters also benefit from shared knowledge~\citep{pfeiffer2023modular}.

\section{Method}
\textbf{Overview.}
Taking inspiration from knowledge transferring between different tasks in multi-task learning, we propose HyperMoE. 
The key idea of HyperMoE is to enhance the availability of knowledge for the current input through positive knowledge transfer between experts.
Through the condition input we designed, the relevant knowledge within the cross-expert information captured by the hypernetwork is encoded into HyperExpert, serving as supplementary information for the currently selected experts. 
In this work, we introduce conditional expert, in which we use shared hypernetworks to generate the expert weights based on the information of the unselected experts. 
The hypernetworks capture information across experts and transfer relevant knowledge to the selected experts by conditional generation.

\subsection{Conditional Expert}

In the transformer model based on the MoE structure, the experts ${\mathnormal{E_1}, E_2, \cdots, E_N}$ in MoE are typically denoted as a group of parallel FFNs. For an input $x\in \mathbb{R}^{d_{\rm{in}}}$, the output $y\in \mathbb{R}^{d_{\rm{out}}}$ can be calculated by the FFN layer as follows:
\begin{equation}
    y=\mathrm{FFN(x)}=\sigma (x\mathrm{W_1})\mathrm{W_2},
\end{equation}
where $\mathrm{W_1}\in \mathbb{R}^{d_{\rm{in}}\times b}$ and $\mathrm{W_2}\in \mathbb{R}^{b\times d_{\rm{out}}}$ are weight matrices. $\sigma(\cdot)$ denotes a non-linear activation function.

In our approach, the matrices $\mathrm{W_1}$ and $\mathrm{W_2}$ are generated by a hypernetwork as described in Section \ref{s32}. In addition, we adopt a bottleneck structure for the conditional expert to improve parameter efficiency inspired by the Adapter~\citep{pmlr-v97-houlsby19a}. Specifically, the bottleneck dimension $b$ satisfies $b\ll d_\mathrm{in / out}$ in our method.

\subsection{HyperExpert}
\label{s32}
These works~\citep{karimi-mahabadi-etal-2021-parameter,he2022hyperprompt,pmlr-v202-phang23a,ivison2023hint} indicate that hypernetworks can learn the parameter information of the main neural network under different input scenarios and efficiently adjust the parameters of the target network to adapt to this information.

Consequently, we propose a novel design called HyperExpert, which captures beneficial knowledge from cross-expert through conditional generation to serve as auxiliary information for the selected experts involved in the computation, as shown in Figure~\ref{fig2}. 
This also results in the extra parameters increasing sub-linearly with the number of layers, enhancing the parameter efficiency of the model.

\noindent\textbf{Selection Embedding.} 
We define the selection embedding to encode the information of experts not selected for each token. Let $p_i \in \mathbb{R}^{d_{\rm{sel}}}$ denote the selection embedding for $i$-th token and $d_{\rm{sel}}$ denotes the dimension. To calculate the selection embedding efficiently and achieve better generalization, we introduce a group of expert embedding $\{s_n\}^N_{n=1}$, where $s_n \in \mathbb{R}^{d_{\rm{exp}}}$ represents the $n$-th expert out of $N$ experts.
The computation process is as follows:
\begin{equation}
    \hat{Z}_i = \mathrm{I} - Z_i=\mathrm{I} - G(x_i),
\end{equation}
\begin{equation}
    p_i=\mathrm{MLP} (\sum_{j=1}^{N} s_j\frac{\hat{z}_{i,j}}{ {\textstyle \sum_{j=1}^{N}} \hat{z}_{i,j}}),
\end{equation}
where $G(\cdot) $ denotes Noisy Top-K Network as described in Section~\ref{s21}. The vector $Z_i \in \mathbb{R}^{|N|}$ represents token-expert allocations: each element $z_{i,j}$ is a binary scalar indicating if the expert embedding $s_j$ is active for the input token $x_i$. $\mathrm{I}$ is an identity vector. $\mathrm{MLP} (\cdot)$ is consisting of two feed-forward layers and a
ReLU non-linearity.

\noindent\textbf{HyperExpert.} 
We use a hypernetwork $H_e(\cdot)$ to construct HyperExpert $\hat{E}$ based on the conditional information of the unselected experts. 
To better share information across different layers and improve parameter efficiency, we \textbf{share the hypernetwork among all layers}. 
Additionally, we define the layer embeddings $l_{\tau}\in \mathbb{R}^{d_{\rm{lay}}}$ for the $\tau$-th Transformer layer. After that, we feed a concatenation of selection embedding and layer embedding to a project network to acquire final embedding $k_{\tau}^{i} \in \mathbb{R}^{d_{\rm{fin}}}$, which is the input to hypernetwork $H_e(\cdot)$ to generates the weight matrices $\mathrm{D}_i^\tau$ and $\mathrm{W}_i^\tau$ for HyperExpert:
\begin{equation}
    (\mathrm{D}_i^\tau, \mathrm{U}_i^\tau) = H_e(k_\tau^i) = (\mathrm{W^D}, \mathrm{W^U})k_\tau^i.
\end{equation}

The hypernetworks $H_e(\cdot)$ consist of a ReLU non-linearity and a feed-forward layer. The weight matrices ${\mathrm{W^D}} \in \mathbb{R} ^{{d_{\rm{fin}}}\times (d_{\rm{in}}\times b)}$ and $\mathrm{{W^U}}\in \mathbb{R} ^{{d_{\rm{fin}}}\times (b\times d_{\rm{out}})}$ are used to generate the down-projection matrix $\mathrm{{D}}_i^{\tau}\in \mathbb{R} ^{d_{\rm{in}}\times b}$ and the up-projection matrix $\mathrm{{U}}_i^{\tau}\in \mathbb{R} ^{b\times d_{\rm{out}}}$ in the HyperExpert $\hat{E}_i$ for $i$-th token at $\tau$-th transformer block.

% The weight matrices of hypernetworks ${W^D}$ and ${W^U}$ are used to generate the down-projection matrix ${D}_i^{\tau}\in \mathbb{R} ^{d\times b}$ and the up-projection matrix ${U}_i^{\tau}\in \mathbb{R} ^{b\times d}$ in the HyperExpert $\hat{E}_i$ for $i$-th token at $\tau$-th transformer block.

Finally, we insert HyperExpert into the expert layer of MoE in parallel and calculate the output of $i$-th token as follows:
\begin{equation}
    \hat{E_i}(x_i) = \mathrm{Relu}(D_i^\tau x_i)U_i^\tau,
\end{equation}
\begin{equation}
    y_i = \sum_{r=1}^{N} G(x_i)E_r(x_i) + \hat{E_i}(x_i).
\end{equation}

In this way, the hypernetwork acts as an information capturer across experts, while the selection embeddings efficiently extract knowledge of experts suitable for the current token selection from the hypernetwork and generate HyperExpert to reduce the transfer of negative knowledge in cross-expert information. The notations and definitions are detailed in Appendix \ref{a3}.
\begin{table*}[ht]
\setlength\tabcolsep{0.3cm}
\centering
\resizebox{0.95\textwidth}{!}{
\begin{tabular}{l|cccccccc>{\columncolor{gray!40}}c}
\toprule 
\multicolumn{10}{c}{\textbf{GLUE}} \\
\midrule
\textbf{Method}  & \textbf{CoLA}& \textbf{SST-2}& \textbf{STS-B}& \textbf{MRPC}& \textbf{QQP}& \textbf{MNLI}& \textbf{QNLI}& \textbf{RTE}& \textbf{Avg} \\
\midrule
{$\text{MoE} $} & {{54.24}}& {{93.81}}& \textbf{{88.69}}& {87.90}& \textbf{{{90.58}}}& {{87.93}}& {91.68}& {67.35}& {82.77}
\\

{$\text{MoE-Share} $}  & {{53.98}}& {{94.27}}& {88.38}& {89.21}& {{90.51}}& {{87.95}}& {92.25}& \textbf{{67.52}}& {83.01}
\\

{$\text{HyperMoE (ours)}$}  & {\textbf{54.67}}& {\textbf{94.38}}& {88.68}& \textbf{{89.63}}& {{90.52}}& {\textbf{88.43}}& \textbf{{92.64}}& {67.01}& \textbf{{83.25}}
\\
\midrule
\multicolumn{10}{c}{\textbf{SuperGLUE}} \\
\midrule
\textbf{Method} &  \textbf{BoolQ}& \textbf{CB}& \textbf{MultiRC}& \textbf{COPA}& \textbf{ReCoRD}& \textbf{RTE}& \textbf{WIC}& \textbf{WSC}& \textbf{Avg} \\
\midrule
{$\text{MoE} $} & {{72.69}}& {{69.64}}& {{66.38}}& {45.00}& {71.26}& {{67.15}}& {63.63}& {56.58}& {64.04}
\\
{$\text{MoE-Share} $} & {{72.11}}& {{67.85}}& {66.71}& {45.00}& {{71.91}}& {\textbf{67.87}}& \textbf{{65.36}}& \textbf{{56.84}}& {64.21}
\\

{$\text{HyperMoE (ours)}$}  & {\textbf{73.14}}& {\textbf{69.68}}& \textbf{{67.68}}& {45.00}& {\textbf{74.06}}& {{67.67}}& {65.31}& {56.53}& \textbf{{64.88}}
\\
\bottomrule
\end{tabular}
}
\caption{\label{t1}
Overall comparison on GLUE and SuperGLUE. {Switch Transformer-base-8} is used as the PLM backbone of all methods. For STS-B, we report Pearson Correlation. For MultiRC, we report F1. For ReCoRD, we report Exact Match. For CoLA, we report Matthews correlation. For other tasks, we report accuracy.
The best result on each block is in \textbf{bold}.
}
\end{table*}

\section{Experiments}
\subsection{Datasets}
We evaluate HyperMoE on 20 NLP datasets across diverse tasks including sequence classification, question answering,  summarization, and text generation. GLUE~\citep{wang-etal-2018-glue} and SuperGLUE~\citep{NEURIPS2019_4496bf24} benchmarks are widely used evaluation datasets for assessing natural language understanding capabilities. Both of them are a collection of text classification tasks: sentence similarity (STS-B;~\citealp{sstb}), (MRPC;~\citealp{mrpc}), (QQP;~\citealp{wang-etal-2018-glue}), question-answering (BoolQ;~\citealp{boolq}), (MultiRC;~\citealp{multirc}), (RECORD;~\citealp{record}), sentiment analysis (SST-2;~\citealp{sst2}), sentence acceptability (CoLA;~\citealp{Cola}), natural language inference (MNLI;~\citealp{mnli}), (QNLI;~\citealp{qnli}), (RTE;~\citealp{rte}), (CB;~\citealp{cb}), word sense disambiguation (WIC;~\citealp{wic}), coreference resolution (WSC;~\citealp{wsc}) and sentence completion (COPA;~\citealp{copa}). For the question-answering task, we consider SQuAD v1.1~\citep{squad}, a collection of question-answer pairs derived from Wikipedia articles, with each answer being a text span from the corresponding reading passage. For the summarization task, we use Xsum~\citep{narayan-etal-2018-dont} and CNN/Daily Mail(CNNDM)~\citep{cnn} to test the model's ability to summarize articles. And finally, the WikiText-2 dataset~\citep{merity2017pointer} is used to measure the ability of long-range dependencies generation.

\subsection{Experiments Details}
\label{s43}
% \vspace{10pt}
Following~\citep{he-etal-2023-merging}, we fine-tune pre-trained MoE models on downstream tasks and report results from the last checkpoint. Unless otherwise specified, Our base model primarily uses Switch Transformer-base-8, which is an MoE model built on T5-base~\citep{JMLR:v21:20-074} with 8 available experts, having a total number of parameters of 620M. 
For the WikiText dataset, we employ GPT-2 small~\citep{radford2019language} as the base model and expand it into the MoE structure by duplicating the weights of the feed-forward layer. In addition, we also use Switch Transformer-base-16/32 to explore the effect of expert numbers on our method. To achieve a fair comparison, all methods in our paper employ the same Top-1 routing and auxiliary loss. For different data scales, we grid-search the training epoch and batch size from \{10, 15, 20\}, and \{8, 16, 32, 64\}, respectively. The learning rate is grid-search from \{1e-5, 5e-5, 1e-4, 5e-4\} with Adam optimizer and the first 10\% warm-up steps. We set the maximum token length to 1024 for WikiText datasets, 348 for SQuAD, and 256 for all other datasets except for the summarization task. For Xsum and CNNDM, we set the max length of source articles to be 1024 and the max length of the target summary to be 128. As for 
All experiments run for 3 times with different seeds and we report the average for each result.

\subsection{Baselines}
Our approach is built upon Switch Transformer~\citep{fedus2022switch}, a well-known MoE model using Top-1 routing. Consequently, we primarily compare our approach with the following baselines: (1) \textbf{MoE}, fully finetuning switch transformer model. (2) \textbf{MoE-Share}, as it is a relevant baseline that does not exploit the inductive bias of the relationship between selected and unselected experts in the process of computation: add an MLP network that is shared among all experts in the MoE layer of a switch transformer, which has the same size as the experts in MoE.

\subsection{Results and Analysis}
\subsubsection{Main Results}

\noindent\textbf{GLUE and SuperGLUE.} 
Table ~\ref{t1} shows the results of various methods applied to the tasks within GLUE and SuperGLUE. Overall, our method improves significantly compared to both MoE and MoE-Share. Specifically, compared to MoE, our method shows a +0.48\% and +0.84\% increase on the GLUE and SuperGLUE benchmarks, respectively. 
This enhancement underscores the advantage of adopting expert knowledge transfer in improving the performance of MoE models.
It's noteworthy that MoE-Share is relevant to ours, but performs worse than MoE on certain datasets such as STS-B, CoLA, and BoolQ. 
A potential reason is that the cross-expert information captured through a shared network cannot achieve effective positive transfer, adversely impacting MoE-Share's effectiveness on these datasets. In contrast, our method maintains a lead on these datasets while also performing well on most other datasets. 
This underscores the effectiveness of our conditional generation strategy: selectively transferring knowledge by leveraging expert selection information during the computation process.

\begin{table}
\centering
\resizebox{0.48\textwidth}{!}{
\begin{tabular}{lcccc}
\toprule
\multirow{2}{*}{{Method}} &\multicolumn{2}{c}{\textbf{Sum.Task}} &\textbf{QA.Task} & \textbf{{Modeling.Task}} \\
\cmidrule(r){2-3}\cmidrule(r){4-4}\cmidrule(r){5-5}
~ & \textbf{{XSum}} & \textbf{CNNDM} & \textbf{SQuAD} & \textbf{WikiText}
\\
\midrule
MoE  & 19.35  &19.75 &83.01 &21.71\\
MoE-Share & 19.41  &19.80 &82.87 &21.63\\
HyperMoE & \textbf{19.67}  &\textbf{20.12} &\textbf{83.51} &\textbf{21.49}\\

\bottomrule
\end{tabular}
}
\caption{\label{t2}
Overall comparison on Xsum, CNNDM, SQuAD, WikiText. For Xsum and CNNDM, we report the Rouge-2 metric~$(\uparrow)$. For SQuAD, we report the Exact Match metric~$(\uparrow)$. For WikiText, we report the Perplexity metric~$(\downarrow)$. All tasks are conducted on the Switch Transformer, except for WikiText, which is carried out on Bert with an MoE structure, as detailed in Section \ref{s43}.
} 

\end{table}
\noindent\textbf{Other Tasks.} 
Table ~\ref{t2} displays the performance of various methods across question-answering tasks, summarization tasks, and text-generation tasks. 
In addition to achieving outstanding performance on Natural Language Understanding (NLU) tasks represented by GLUE and SuperGLUE, our method also excels in Natural Language Generation (NLG) tasks. 
Experimental results show that our method outperforms baseline methods across all NLG tasks. 
Specifically, in extractive question-answering tasks, our method shows improvements of 0.50\% and 0.64\% over MoE and MoE-Share, respectively. 
Like the NLU tasks, MoE-Share again underperforms, indicating that the extra networks may not effectively learn information useful to experts without the expert selection inductive bias.
Furthermore, our method still performs well in summarization tasks involving long-text inputs. 
This demonstrates that our method can still effectively enhance the availability of expert knowledge through knowledge transfer under complex input conditions, suitable for tasks of various text lengths.
Lastly, our method also achieves considerable improvement on Wikitext.
These results demonstrate the effectiveness of HyperMoE in various tasks.

Additionally, we compared our method with a related baseline~\citep{do2023hyperrouter} that uses a hypernetwork to generate conditional parameters for the routing of MoEs, aiming to address the routing allocation problem in SMoE methods, where all experts tend to have similar representations. Experimental results can be found in Appendix~\ref{b1}.

\subsubsection{Ablation Study}
We conduct an ablation study on the SQuAD to evaluate the effectiveness of the proposed modules. 
The embedding design is removed to verify the effect of using external information as embeddings.
As shown in Table~\ref{t3} (row 1), when the embedding and hypernet are removed, our method is equivalent to MoE. 
Table~\ref{t3} (row 2) omits the embedding design, directly using the sample's hidden state as input to the hypernetwork. 
This results in a marked decrease in performance, even falling below that of MoE.
This suggests that conditioning the hypernetwork on the sample enlarges the parameter search space and is difficult to optimize.
In an additional experiment, we use a depthwise separable convolutional network~\citep{howard2017mobilenets} with kernels of sizes 5×5 and 3×3 to compress and reduce the dimensions of the experts' weights, obtaining expert embeddings. More details are in Appendix~\ref{a1}. The selection embeddings are then computed and input into the hypernetwork as described in Section~\ref{s32}.
Empirically, expert weights can better represent the information of experts. However, as shown in Table~\ref{t3} (row 3), this strategy leads to a slight drop in performance, defying expectations. 
A potential explanation is the substantial information loss associated with compressing expert weights, resulting in a loss of specific information details. We leave the exploration of this strategy to future work.

\begin{table}
\centering
\resizebox{0.45\textwidth}{!}{
\begin{tabular}{cc|c}
\toprule
\textbf{Embedding}&\textbf{Hypernet} &\textbf{Exact Match}  \\
\midrule
\XSolidBrush & \XSolidBrush  & 83.01  \\
\XSolidBrush & \Checkmark & 82.92  \\
\textbf{W} & \Checkmark & {83.33}  \\
\textbf{P} & \Checkmark & \textbf{{83.51}}  \\

\bottomrule
\end{tabular}
}
\caption{\label{t3}
Ablation study on SQuAD. \textbf{W} represents the use of expert weights as embeddings. \textbf{P} denotes the use of our proposed selection embedding.
}

\end{table}

\begin{figure}
\centering
\includegraphics[width=1.0\columnwidth]{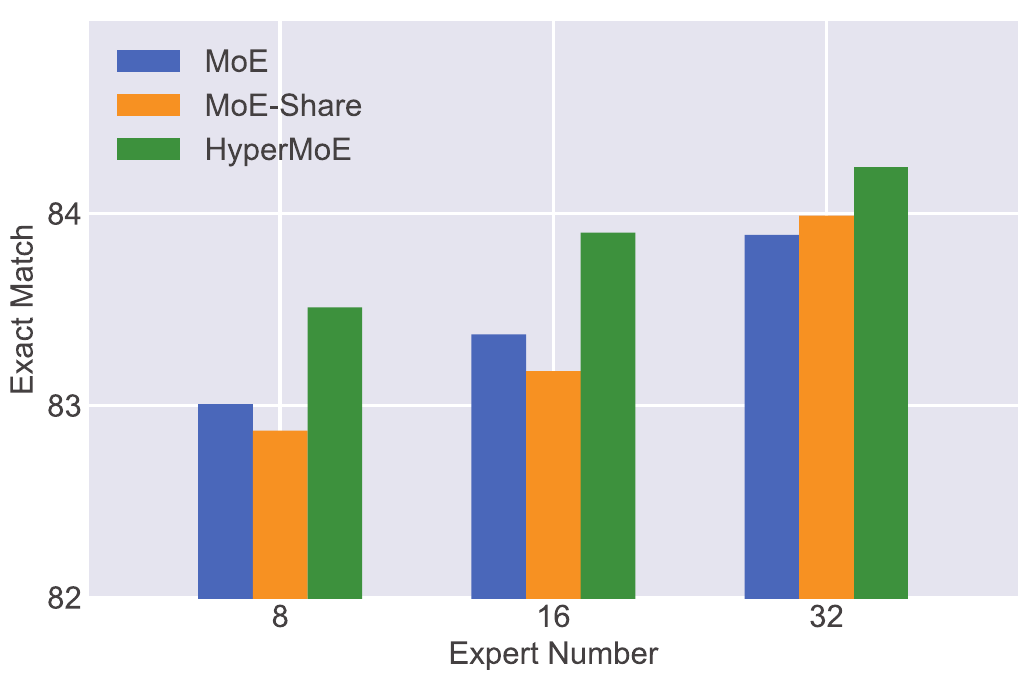}

\caption{Performance comparison of MoE methods on the SQuAD dataset with the increase in the number of experts.  }
\label{fig3}
\end{figure}

\subsubsection{Performance in Scaling the Number of Experts and Model Size}
To explore the impact of the variation in the number of experts on our method, we fine-tuned on the SQuAD dataset using Switch Transformer-base-16/32 as pre-trained models. 
These models possess 16 and 32 experts in each MoE layer, respectively. 
As demonstrated in Figure~\ref{fig3}, every method achieves performance enhancement across models featuring a diverse number of experts. 
Notably, our method exhibits consistent superior growth and outperforms the others. 
This indicates that the proposed conditional generation strategy can still effectively benefit from knowledge transfer as the number of experts increases.

\begin{figure}[ht]
\centering
\includegraphics[width=1.0\columnwidth]{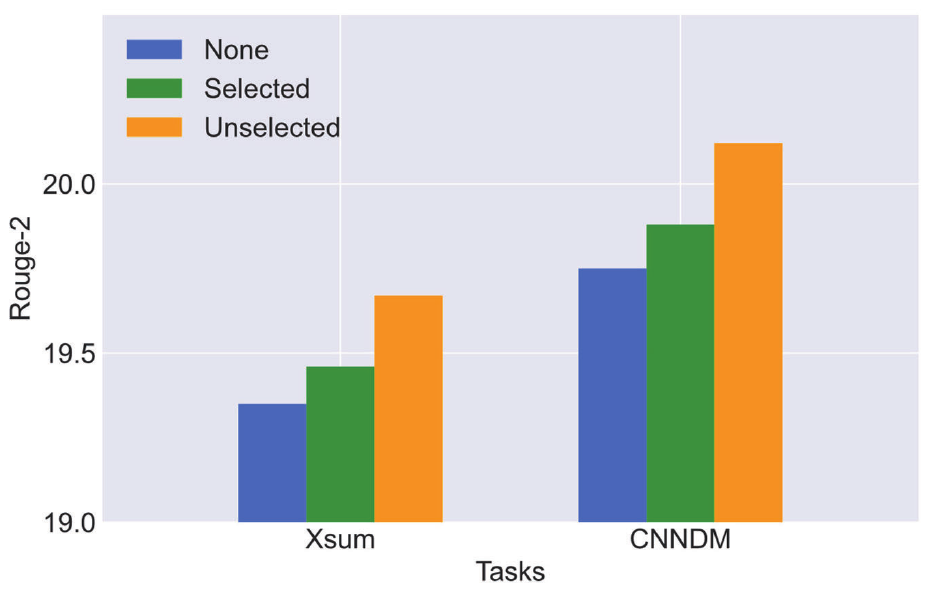}

\caption{Compare the performance of our method when calculating selection embedding using the selected expert embeddings or the unselected expert embeddings, respectively.}
\label{fig4}
\end{figure}

In addition, to verify the effectiveness of our method on larger parameter-scale language models, we fine-tuned DeepSeek-MoE~\citep{dai2024deepseekmoe}, which has 16 billion parameters, on the Open-Platypus~\citep{lee2023platypus} dataset and tested the fine-tuned model on MMLU~\citep{mmlu}, GSM8K~\citep{gsm8k}, and BBH~\citep{bbh}. As shown in Table~\ref{t7}, our method consistently achieved improvements. This indicates that our method can be widely applied to language models ranging from hundreds of millions to tens of billions of parameters.

\begin{table}
\centering
\resizebox{0.40\textwidth}{!}{
\begin{tabular}{c|ccc}
\toprule
\textbf{Method}&\textbf{MMLU} &\textbf{GSM8K} &\textbf{BBH} \\
\midrule
MoE & 46.52  & 17.97 & 39.38   \\
HyperMoE & \textbf{47.06} & \textbf{18.75} &\textbf{40.77}   \\

\bottomrule
\end{tabular}
}
\caption{\label{t7}
Comparison of results after fine-tuning on DeepSeek-MoE 16B.
}

\end{table}

\subsubsection{Investigating of Selection Embedding.}
\noindent\textbf{The unselected expert embeddings are more informative than selected expert embeddings.} Empirically, by conditioning on the information of unselected experts, specific knowledge can be extracted from cross-expert knowledge, which selected experts do not possess, thereby aiding the selected experts.
To verify this idea, we input embeddings of both selected and unselected experts into a hypernetwork, comparing their performance on the Xsum and CNNDM datasets. As shown in Figure~\ref{fig4}, using unselected expert information as conditional input can achieve comparable results. 
This implies that the conditional information of unselected experts can generate more beneficial knowledge for the selected experts through a shared hypernetwork.

\begin{figure}[ht]
\centering
\includegraphics[width=1.0\columnwidth]{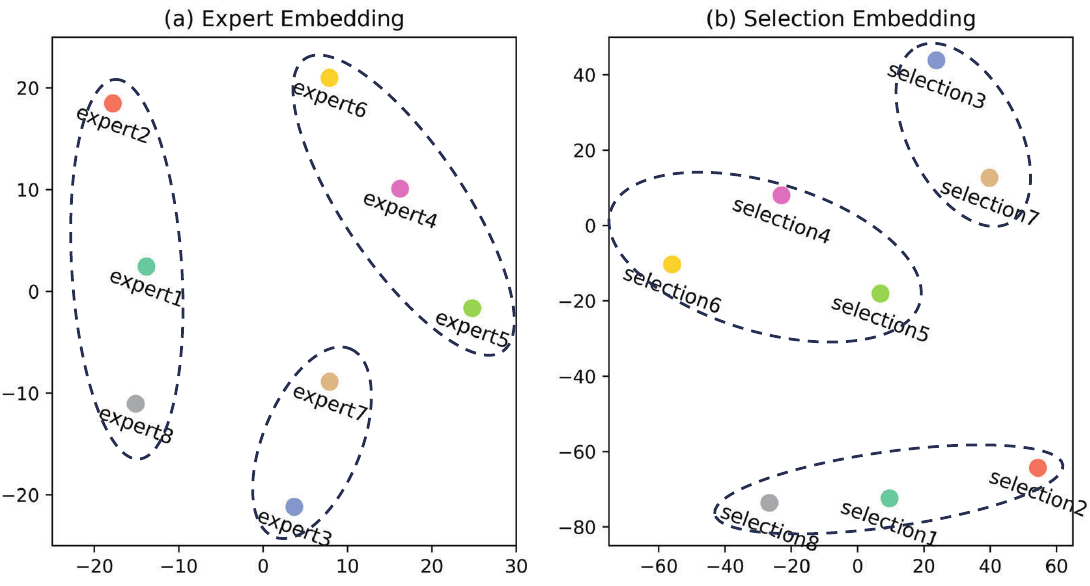}

\caption{t-SNE visualizations for expert embeddings (right) and selection embeddings (left). selection $i$ denotes calculated using all expert embeddings except for the $i$-th expert embedding. }
\label{fig5}
\end{figure}

\noindent\textbf{Expert embeddings and the selection embeddings have a corresponding relationship.}
In addition, to explore whether the embeddings encode the information in our proposed method, we provide visualizations of the expert embeddings and computed selection embeddings within the final MoE layer of Switch Transformer-base-8 learned on CNNDM. 
Figure~\ref{fig5} reveals that both sets of embeddings exhibit sparse distributions, suggesting that the embeddings encode some specific non-relevant information. 
We also observe a correlation between the distances among selection embeddings and the distances among expert embeddings, such as between 4-5-6, 1-2-8, 7-3. 
This correlation implies that the information of the unselected experts encoded by the selection embeddings depends on the information of the selected experts, further illustrating that the selection embeddings effectively capture the information of the knowledge the currently selected experts need.

\subsubsection{Impact of Additional Computation}
Although our method achieves significant performance improvements compared to the original MoE structure, it introduces additional networks, which inevitably slightly reduces the inference speed of HyperMoE. We evaluate the number of samples per second that our method can train/infer based on Switch Transformer-base-8. The methods in each task employ the same batch size.
As shown in Table~\ref{t4}, our method's training/inference speed is only reduced by about 15\% and 10\% compared to MoE, respectively. This suggests that our approach can enhance the availability of expert knowledge more effectively without significantly increasing computational costs while maintaining sparsity during expert selection.

\begin{table}
\centering
\resizebox{0.45\textwidth}{!}{
\begin{tabular}{lccc}
\toprule
\multirow{2}{*}{{Method}} & \multicolumn{2}{c}{\textbf{Sum.Task}} &\textbf{QA.Task}  \\
\cmidrule(r){2-3}\cmidrule(r){4-4}
~ & \textbf{{XSum}} & \textbf{CNNDM} & \textbf{SQuAD} 
\\
\midrule
MoE$_{train}$  & 76.31  &77.20 &89.56\\
HyperMoE$_{train}$ & {65.69}  &65.95 &{75.38} \\
\midrule
MoE$_{eval}$  & 5.42  &5.99 &78.43\\
HyperMoE$_{eval}$ & {4.78}  &5.32 &{67.73} \\
\bottomrule
\end{tabular}
}
\caption{\label{t4}
The number of samples trained/evaluated per second.
}

\end{table}

\section{Related Work}
\subsection{Mixture of Expert}
% In the field of NLP, 
\citet{shazeer2017} introduces Mixture-of-Expert layers for LSTM language modeling and machine translation. 
These layers are inserted between the standard layers of the LSTM model. Subsequent work primarily builds on Transformers, where expert layers often replace dense layers. 
\citet{lepikhin2021gshard} first introduces MoE layers into Transformers and studies them in the context of machine translation. With the release of Gshard~\citep{lepikhin2021gshard} and Switch Transformer~\citep{fedus2022switch}, MoE models are scaled up to new heights by introducing thousands of small-scale experts. 
In terms of routing, \citet{shazeer2017} use routing to the top k experts out of k > 1. \citet{hazimeh2021dselect} propose DSelect-k, a smoothed version of the top-k routing algorithm that improves upon standard top-k routing. 
\citet{fedus2022switch}, \citet{pmlr-v162-clark22a} and \citet{openmoe2023} demonstrate that top-1 routing can also achieve competitive results. 
Hash Layer~\citep{roller2021hash} and StableMoE~\citep{dai2022stablemoe} employ fixed routing strategies for more stable routing and training. 
\citet{zhou2022mixture} propose an expert selection routing strategy where each token can be assigned to a different number of experts. \citet{qiu2023emergent} demonstrates that a standard LM can be fine-tuned as its Mix of Experts (MoE) counterpart, effectively improving downstream in-domain and out-of-domain generalization capabilities.
\citet{pmlr-v162-rajbhandari22a} and \citet{dai2024deepseekmoe} isolate general knowledge from experts using shared experts from engineering and algorithm perspectives, respectively, to promote expert specialization. 

In contrast to previous work, our work mainly focuses on the knowledge transfer between experts in MoE. 
This provides a solution for improving the availability of expert knowledge in MoE while maintaining sparsity.

\subsection{HyperNetwork}
Hypernetworks~\citep{ha2017hypernetworks} are widely used in multi-task learning due to their ability to avoid negative interference of corresponding modules by soft parameter sharing and generating module parameters conditioned on the shared parameters. 
The most common approach usually takes task~\citep{karimi-mahabadi-etal-2021-parameter,zhao-etal-2023-prototype} or language embeddings~\citep{ustun2020udapter,baziotis2022multilingual} as contextual information to generate corresponding module parameters, such as adapter layers~\citep{ustun2020udapter,ansell2021mad,karimi-mahabadi-etal-2021-parameter}, classifier heads~\citep{ponti-etal-2021-parameter}, and continuous prompts~\citep{he2022hyperprompt}. In addition, hypernetwork-based approaches have also been very successful in zero-shot and few-shot scenarios~\citep{deb2022boosting,pmlr-v202-phang23a,ivison2023hint}. 

In the field of NLP, hypernetworks are mainly used to improve the generalization~\citep{volk2022example,zhang2023hypernetwork} and applicability~\citep{wullach2022character,he2022hyperprompt,tan2023massive} of dense models. HyperRouter~\citep{do2023hyperrouter} chooses to solve the routing allocation problem in SMoE methods by using a fixed HyperNetwork to generate routing weights.
Our work explores the integration of hypernetworks with sparse MoE. 
We propose to input the expert selection status of tokens as information into the hypernetwork and generate module parameters that correspond to the respective tokens. 
To the best of our knowledge, this is the first time that hypernetworks have been introduced in the MoE structure, this is the first time HyperNetwork has been applied to the expert structure in MoEs, introducing inductive bias through the hypernetwork: information from unselected experts is reused, which extends the application scope of hypernetworks and provides new insights for knowledge transferring in MoE.

\section{Conclusion}

In this work, we introduce HyperMoE, a novel Mixture of Experts (MoE) architecture. Inspired by the concept of knowledge transfer in multi-task learning, we propose a method to facilitate knowledge transfer between experts through conditional generation. 
Our method enhances expert knowledge availability while maintaining expert selection's sparsity.
We show the effectiveness of our approach across a wide range of NLP tasks. Experimental results demonstrate that our method exhibits excellent performance compared to the conventional MoE. 
Furthermore, our analysis shows that without any measures, there could be negative knowledge transfer across experts when transferring knowledge to specific experts. 
Our approach mitigates this issue by capturing the contextual information of experts. 
We explore the feasibility of knowledge transfer between experts in MoE, providing a new perspective for future improvements in MoE architectures.

\section*{Limitations}
Despite our work has demonstrated strong experimental results, there are several limitations: (1) In this work, we utilize end-to-end training to learn expert embeddings. Incorporating prior knowledge, such as expert weights, into the embedding learning process may improve efficiency and performance. We will improve upon this in future work. 
(2) We insert HyperExpert into the expert layer of MoE in parallel. This incurs additional computational overhead. Mitigating this issue could be achieved by employing some parameter-efficient methods (such as LoRA~\citep{hu2022lora} and (IA)$^3$~\citep{liu2022few}) to insert HyperExpert into MoE. 
(3) Current experiments mainly focus on fine-tuning the pre-trained MoE model. Utilizing our proposed method to train a large-scale MoE from scratch will be the emphasis of our future work.

\section*{Acknowledgements}
This work is supported by the National Key R\&D Program of China (Grant No. 2021ZD0110100), National Natural Science Foundation of China (Grant No. 62176025, U21B2045), Beijing Nova Program (Grant No.20220484161), and Theme-based Research Scheme (T45-205/21-N) from Research Grants Council of Hong Kong and Development Centre from InnoHK.

% Bibliography entries for the entire Anthology, followed by custom entries
%\bibliography{anthology,custom}
% Custom bibliography entries only

\bibliography{anthology,custom}
\bibstyle{acl_natbib}
\clearpage
\appendix

\section{Depthwise Separable Convolutional Networks Details}
\label{a1}

For every expert weight in each MoE layer, we use the same convolutional network to reduce its dimensionality. First, we stack them so that their dimensional form is three-dimensional, similar to images. Then, we perform convolution on them. Our experiments used depthwise separable convolutions, with specific parameters as shown in Table~\ref{t5}.

\begin{table}[hp]
\centering
\resizebox{0.48\textwidth}{!}{
\begin{tabular}{c|c|c}
\toprule
\textbf{Type/Stride}&\textbf{Filter Shape} &\textbf{Input Size}  \\
\midrule
Conv dw / s5 & $5\times5\times2$ dw  & $8\times2\times3072\times768$  \\
Conv / s1 & $1\times1\times2\times32$ & $8\times2\times614\times153$  \\
Avg Pool / s(16, 6)& Pool(16,6)  & $8\times32\times614\times153$\\
Conv dw / s3 & $3\times3\times32$ dw  & $8\times32\times38\times25$  \\
Conv / s1 & $1\times1\times32\times128$ & $8\times32\times12\times8$  \\
Avg Pool / s8& Pool(8,8)& $8\times128\times12\times8$ \\
Output & -- & $8\times128\times1\times1$ \\

\bottomrule
\end{tabular}
}
\caption{\label{t5}
Specific parameters and structure of depthwise separable convolutions.
}

\end{table}
The compressed expert weights are used as expert embeddings in subsequent computations as described in Section~\ref{s32}.

\section{Additional Results}
\label{b1}
Here, We conducted additional experiments to demonstrate the superiority of our method over similar methods. The experiment is based on SwitchTransformer and GPT2-MoE, with all methods having the same number of parameters and experts. Moreover, we trained GPT2-MoE from scratch on WikiText-103, and the experimental results are as shown in Table~\ref{t6}.

\begin{table}
\centering
\resizebox{0.48\textwidth}{!}{
\begin{tabular}{c|cccc}
\toprule
\textbf{Method}&\textbf{SQuAD} &\textbf{CNN} &\textbf{WikiText-2} &\textbf{\makecell{WikiText-103\\(from scratch)}} \\
\midrule
MoE & 83.01  & 19.75 & 21.71 & 22.0  \\
HyperRouter & 78.77 & 15.56 & 21.69 & 22.21  \\
HyperMoE & \textbf{83.51} & \textbf{20.21} &\textbf{21.49} & \textbf{21.81}  \\

\bottomrule
\end{tabular}
}
\caption{\label{t6}
Results of the comparison with HyperRouter on 4 datasets(SQuAD, CNN, WikiText-2, WikiText-103). We conduct pre-training experiments on WikiText-103, while the other three datasets are used for fine-tuning experiments.
}

\end{table}

The results show that our method achieves the best performance both scratch training and finetuning. The way HyperRouter generates router weights causes the model to lose its original pre-trained weights, making HyperRouter ineffective in fine-tuning scenarios. The reason HyperRouter's performance is inconsistent with the original results under the scratch training scenario might be because we fixed the number of experts chosen during the training process, instead of gradually increasing the number of chosen experts during training as in the original setup. In the original setting, the computational cost of HyperRouter is significantly higher than the method with a fixed number of experts. For a fair comparison, we chose the same expert selection setting. Experimental results show that HyperMoE can efficiently transfer knowledge between different experts in MoE when choosing the same number of experts.

\section{Table for notations}
\label{a3}

\begin{table}[hp]
\centering
\resizebox{0.48\textwidth}{!}{
\begin{tabular}{c|c}
\toprule
\textbf{Variable}&\textbf{Definition}  \\
\midrule
$N$& the number of experts in MoE    \\
$E(\cdot)$ & a typical expert layer   \\
$G(\cdot)$& the noisy top-k network  \\
$x_i$& the $i$-th input token \\
$d_{\mathrm{in}},d_{\mathrm{out}}$ & hidden dimension of the model    \\
$b$ & dimension of bottleneck   \\
$\sigma(\cdot)$&  non-linear activation function \\
$p_i$ & selection embedding for $i$-th token \\
% $t$ & dimension of selection embedding \\
$d_{\mathrm{sel}}$ & dimension of selection embedding \\
$s_n$ & the $n$-th expert embedding out of $N$ experts\\
% $m$ & dimension of expert embedding \\
$d_{\mathrm{exp}}$ & dimension of expert embedding \\
$z_{i,j}$ & a binary scalar indicating if $q_j$ is active for $x_i$\\
% $H_e(\cdot)$ & the hypernetwork consist of a ReLU layer and a feed-forward layer \\
$l_{\tau}$ & layer embedding for the $\tau$-th Transformer layer\\
% $t'$ & dimension of layer embedding \\
$d_{\mathrm{lay}}$ & dimension of layer embedding \\
$k$ & final embedding as input to hypernetwork\\
% $k$ & dimension of final embedding \\
$d_{\mathrm{fin}}$ & dimension of final embedding \\
$\mathrm{W^D}$ & weight matrices of hypernetwork\\
$\mathrm{W^U}$ & weight matrices of hypernetwork \\
$\mathrm{D}$ & down-projection matrix of HyperExpert \\
$\mathrm{U}$ & up-projection matrix of HyperExpert \\

\bottomrule
\end{tabular}
}
\caption{\label{t6}
The notations and definitions.
}

\end{table}

\end{document}